\documentclass[pdflatex,sn-mathphys-num]{sn-jnl}
\usepackage{array,multirow,graphicx}
\usepackage{amsmath,amssymb,amsfonts}%
\usepackage{amsthm}%
\usepackage{mathrsfs}%
\usepackage[title]{appendix}%
\usepackage{xcolor}%
\usepackage{natbib}%
\usepackage{babel}%
\usepackage{textcomp}%
\usepackage{manyfoot}%
\usepackage{booktabs}%
\usepackage{listings}%
\usepackage{geometry}
\usepackage{tikz}
\usetikzlibrary{shapes.misc}
\usepackage{tabularx}

\theoremstyle{thmstyleone}%
\theoremstyle{thmstyletwo}%

\theoremstyle{thmstylethree}%

\begin{document}
\newcommand{\STAB}[1]{\begin{tabular}{@{}c@{}}#1\end{tabular}}

\title[Article Title]{A Survey of Large Language Models for European Languages}

%%=============================================================%%
%% GivenName	-> \fnm{Joergen W.}
%% Particle	-> \spfx{van der} -> surname prefix
%% FamilyName	-> \sur{Ploeg}
%% Suffix	-> \sfx{IV}
%% \author*[1,2]{\fnm{Joergen W.} \spfx{van der} \sur{Ploeg} 
%%  \sfx{IV}}\email{iauthor@gmail.com}
%%=============================================================%%

\author[a,b]{\fnm{Wazir} \sur{Ali}{\email{wazir.ali@utu.fi}}}
\author[b]{\fnm{Sampo} \sur{Pyysalo}{\email{sampo.pyysalo@utu.fi}}}

\affil[a]{\orgdiv{TurkuNLP Group}, \orgname{The University of Turku}, \postcode{FI-20014}, \country{Finland}}
\affil[b]{\orgdiv{Institute of Business Management}, \postcode{75190}, \country{Pakistan}}

%\orgaddress{\street{} \city{} \state{},
%\affil[2]{\orgdiv{Department}, \orgname{Organization}, \orgaddress{\street{Street}, \city{City}, \postcode{10587}, \state{State}, \country{Country}}}

%\affil[3]{\orgdiv{Department}, \orgname{Organization}, \orgaddress{\street{Street}, \city{City}, \postcode{610101}, \state{State}, \country{Country}}}

%%==================================%%
%% Sample for unstructured abstract %%
%%==================================%%

\abstract{Large Language Models (LLMs) have gained significant attention due to their high performance on a wide range of natural language tasks since the release of ChatGPT. The LLMs learn to understand and generate language by training billions of model parameters on vast volumes of text data. Despite being a relatively new field, LLM research is rapidly advancing in various directions. In this paper, we present an overview of LLM families, including LLaMA, PaLM, GPT, and MoE, and the methods developed to create and enhance LLMs for official European Union (EU) languages. We provide a comprehensive summary of common monolingual and multilingual datasets used for pretraining LLMs. 

%Finally, we discuss open challenges and suggest future research directions. 
%number of language models
%number of datasets and their size
}


\maketitle

\section{Introduction}\label{sec1}
Language modeling has a long-standing history, evolving from rule-based to statistical models \citep{shannon1951prediction}. Early language models (LMs) relied on manually created syntactic and grammatical rules. However, statistical models, which predict the next word based on the preceding ones, analyze real-world text data to understand word co-occurrence patterns. Since then, language models have become essential to many tasks in Natural Language Processing (NLP), including information retrieval \citep{baeza1999modern}, speech recognition \citep{jelinek1998statistical}, and other applications \citep{manning1999foundations}.

Language modeling has progressed from statistical to neural approaches, and subsequently from pretrained LMs to large language models (LLMs). Statistical LMs \citep{ChenG99} treat text as a sequence of words, estimating word probabilities as the product of their conditional probabilities. A popular approach within statistical LMs is the use of Markov models, particularly n-gram models \citep{IyerO99}. However, these models struggle to capture the full variability and diversity of language due to the problem of data sparsity.

Early neural LMs \citep{bengio2000neural, mikolov2010recurrent} addressed the data sparsity problem by mapping words to low-dimensional embeddings, but such models were limited to task-specific data. Unlike early neural LMs, pretrained LMs introduced the paradigm of pretraining and fine-tuning using recurrent neural networks \citep{mikolov2010recurrent} for general NLP tasks. Pretrained LMs outperform traditional language models in terms of performance after fine-tuning for downstream tasks. While pretrained LMs are trained in a self-supervised manner on large corpora to learn generic representations, traditional LMs train task-specific models in a supervised manner. Larger pretrained LMs have led to the development of LLMs by significantly increasing the size of the training datasets (many gigabytes to terabytes) and the number of parameters (billions to trillions).

The success of transformers in language modeling stems from several key factors. Firstly, parallelization allows transformers to scale effectively with massive datasets and train LLMs faster compared to sequential models \citep{megatron-turing-nlg}. Secondly, they handle variable input lengths efficiently \citep{surveyllms}. Thirdly, transformers can capture long-range dependencies, understanding the relationships between words to grasp grammatical structure and meaning. Fourthly, transformers are highly scalable, allowing the use of large datasets, the addition of more layers, and an increase in the number of neurons \citep{GLaM, CPM-2}. Unlike traditional LMs, transformer-based LLMs are not task-specific and can be adapted for a wide range of NLP applications by adding specific output layers on top of the core transformer architecture. However, the generalization capability of LLMs remains an ongoing challenge \citep{Generalizationllms}.

Recent advancements in transformer-based language models pretrained on massive web corpora in a self-supervised setting include Megatron-Turing \citep{megatron2022}, GPT-4 \citep{GPT4}, Bloom-176B \citep{BLOOM-22}, LLaMA-3 \citep{llama3}, Mixtral \citep{MoE}, and the most recent models, Gemini and ChatGPT. These models not only provide state-of-the-art performance on NLP tasks but have also become general task solvers. LLMs outperform traditional pretrained LMs in terms of performance gains in downstream NLP tasks due to their greater capabilities. Many LLMs have been proposed in the literature since this breakthrough.

Much of the research on the development of LLMs is based on the English language. More recent survey articles on LLMs \citep{min2023recent, minaee2024large, overview-llms-24}, training datasets \citep{Datasets-llms-24}, and evaluation methods \citep{survey-evaluation-llms23} present comprehensive overviews of transformer architectures, pretraining, and fine-tuning datasets. In contrast to these surveys, our contribution focuses on providing an overview of the existing work on the development of LLMs for European languages. European languages can be categorized into high-resource, mid-resource, and low-resource languages. To the best of our knowledge, this is the first article that presents a comprehensive review of the existing work on the resources and development of LLMs for official European languages. 

% Our main contributions are summarized as follows.

% \begin{itemize}
%     \item We present a comprehensive survey on the available monolingual and multilingual datasets for training LLMs.
%     \item The summary of LLMs is presented that include major findings, architecture design and training details.
%     \item We summarize in recent progress in  LLMs to help the practitioners and researchers. 
%      \item The article covers the official European languages to present the background, pretraining datasets and the future research. 
% \end{itemize}

\section{Large Language Models}\label{sec2:Background}
This section provides an overview of small and large language models, including encoder, decoder, encoder-decoder, and Sparse models along with a discussion of remarkable findings that have contributed to enhance the performance in NLU and NLG tasks. Table \ref{tab:overview_llms_relatedwork} presents an overview of LLMs, categorized as encoder-only, decoder-only, and encoder-decoder based models. 

\subsection{Encoder-Only Models}
These models solely consists of an encoder network. Initially, these models were created to predict a class label for a given input text in language understanding tasks like text classification. Notable models include BERT~\citep{BERT-19} (Bidirectional Encoder Representations from Transformers), and its variants XLNet~\citep{XLNet}, DistilBERT~\citep{DistilBERT}, ELECTRA~\citep{ELECTRA}, RoBERTa~\cite{RoBERTa}, ALBERT~\citep{albert-20}, DeBERTa~\citep{Deberta} as described below: 

BERT~\citep{BERT-19} is the most widely used encoder-transformer models is bidirectional, meaning it's trained to understand the context of words in a sentence by looking at the words that come before and after it. The architecture of BERT model consists of three modules: i). an embedding module which creates a series of embedding vectors from the input text. ii). a stack of Transformer encoders which create contextual representation vectors from the embedding vectors.
iii). a fully connected layer which creates one-hot vectors from the representation vectors. Moreover, BERT is trained using two main objectives: i). Masked language modeling (MLM) involves predicting masked words in a sentence based on the surrounding context and ii). next sentence prediction aims to predict whether a second sentence follows logically from the first sentence. Following the success of BERT, several researchers have proposed similar encoder-only models.

RoBERTa~\citep{RoBERTa} significantly improves the robustness of BERT using a set of model design choices and training strategies.
ALBERT~\citep{albert-20} adopts a parameter-reduction approach for lower memory consumption and increased training speed compared to the original BERT. DeBERTa~\citep{Deberta} is a decoding-enhanced BERT with disentangled attention, achieves better performance than both BERT and RoBERTa, by employing disentangled attention and an enhanced decoder.
ELECTRA~\citep{ELECTRA} uses Replaced Token Detection (RTD) during pre-training, achieving better results and proving more efficient than MLM.
Later, advancements led to the development of XLMs and encoder-only models that leverage the advantages of autoregressive models for training and inference. XLMs~\citep{xlm} extend BERT to a cross-lingual model using unsupervised and supervised methods, achieving state-of-the-art (SOTA) results on cross-lingual classification and supervised and unsupervised tasks. Furthermore, both XLNet~\citep{XLNet} and UniLM~\citep{UniLM} are encoder-only models which incorporate decoder aspects into their training. XLNet is trained using a generalized autoregressive method, while UniLM uses the combination of uni-directional, bi-directional, and sequence-to-sequence prediction which is useful for both NLU and generation tasks..

\subsection{Decoder-Only Models}
The release of OpenAI's GPT-1~\citep{gpt1} is widely regarded as a pioneer in decoder-only transformer models. By generatively pre-training an LLM on a large unlabeled text and then discriminatively fine-tuning on each individual task. GPT-1 achieved impressive results in text generation, QA, document classification, and semantic similarity assessment. The decoder-only models use Transformer decoders to predict the next word in a sequence based on the previous words unlike encoder-decoder models that process input text and then generate a response. Such models are well-suited for tasks like QA and text generation due to their autoregressive nature. The popular LLMs can be broadly categorised into  a series of GPT-based~\citep{gpt1}, LLaMA-based~\citep{LLaMA}, PaLM-based~\citep{PaLM}, and others~\citep{grok1, megatron-turing-nlg,BLOOM-22,Falcon} described below: 

GPT-1~\citep{gpt1} is the first decoder only powerful Transformer model introduced by OpenAI, trained on large unlabeled data for language understanding and specific tasks.  After training phase, supervised fine-tuning on labeled datasets for specific tasks adapts the pretrained model for the target task. This work demonstrates that Transformer model can achieve SOTA performance on several benchmarks using unsupervised pretraining and task-specific fine-tuning. The limitations included the inefficiency to understand large context in long sequences and a tendency to generate absurd text.  However, GPT-1 was a stepping stone for more advanced decoder-only models despite these limitations. Later, this work was followed at large scale to achieve impressive results in text generation tasks.

GPT-2~\citep{gpt2} was a pioneer in creative text generation with its 1.5 billion parameters on a just 40GB text dataset. However, GPT-3 is similar to GPT-1 architecture with dense and sparse attention in transformer layers. It demonstrates that LLMs can tin on larger batch sizes with lower learning rates. Overall, GPT-3 increases model parameters to 175B, indicating that the performance of LLMs improves with scale and can compete with fine-tuned models. This vast size and richer training allow GPT-3 to not only create impressive creative text formats, but also handle complex tasks like QA, MT, which make it a more versatile and powerful LLM, though still susceptible to biases from its training data.

Similar to GPT-3 transformer, OpenAI introduce GPT-4~\citep{GPT4} by using post-training process which improves factual accuracy and aligns the model with desired behaviors. Moreover, optimization methods that work consistently across different scales make it more robust over precedent models. GPT-4 is more powerful LLM with the ability to handle both text and image inputs to generate text outputs. It achieves human-level results on various benchmarks, even passing a simulated bar exam with a top 10\% score.

PanGu-{\(\alpha\)}~\citep{PanGu} is an autoregressive language model trained specifically for the Chinese language. Inspired by GPT-3 and their preliminary experiments, PanGu-{\(\alpha\)} builds upon the Transformer-based architecture as its foundation. It further enhances this base by adding a custom query layer on top of the Transformer layers. This layer helps guide the model towards the desired output during the pretraining phase. During training, PanGu-{\(\alpha\)} has the ability to handle the huge workload by utilizing MindSpore Auto-parallel. This strategy distributes computations across processors in five different ways. PanGu-{\(\alpha\)} is trained on a large, high-quality Chinese dataset, ensuring its ability to handle diverse tasks. The model's capabilities are evaluated across various NLP tasks, including question answering (QA) and text summarization.

ERNIE 3.0~\citep{ERNIE3.0} leverages multi-task learning with a modular architecture based on Transformer-XL~\citep{Transformer-XL}. It focuses on the Chinese language and claims to achieve SOTA performance on a wide range of NLP tasks. Later, extends to ERNIE 3.0 TITAN~\citep{ERNIE3.0-Titan}, a larger version of ERNIE 3.0 with around 26 times more parameters. It achieves better performance on various NLP tasks and introduces a "Credible and Controllable Generations" objective to enhance consistency in fact in text generation.

Similarly, GPT-NeoX-20B~\citep{GPT-NeoX} closely follows the GPT-3 architecture with some modifications for efficiency. BLOOM~\citep{BLOOM-22} is another example, leveraging a causal decoder model trained on a massive dataset. Efficiency and Novel Techniques While scaling models brings performance gains, it also presents challenges related to computational cost and training difficulty.  Several models address these challenges by incorporating efficiency-enhancing techniques:

MT-NLG~\citep{megatron-turing-nlg} is a large decoder model explores filtered high-quality data and blends various datasets for training, achieving better performance than GPT-3.

Chinchilla~\citep{chinchilla22} model presents a balanced relationship between model size and training data, identifying that doubling the model size requires doubling the training tokens. 

Pathways Language Model (PaLM)~\citep{PaLM} employs parallel attention and feed-forward layers for faster training. It also introduces techniques like SwiGLU activation, RoPE embeddings, and multi-query attention to further enhance efficiency. PaLM-2~\citep{palm2} offers a smaller and more efficient variant of PaLM, while U-PaLM~\citep{u-palm} explores training with an additional objective to improve performance on various tasks.

GLM-130B~\citep{glam30b} is trained on English and Chinese languages with a bidirectional fashion. In contrast to original GLaM~\citep{GLaM} model, GLM-130B incorporates a small amount (5\%) of multi-task instruction data alongside the primary self-supervised mask infilling objective to enhance performance and utilizes an embedding layer gradient shrinkage technique to ensure stability in the training. 

LLaMA~\citep{LLaMA} series of decoder-only LLMs, ranging from 7 to 70 billion parameters, is renowned for its parameters and instruction tuning capabilities. LLaMA-1~\citep{LLaMA} implements efficient causal attention by eliminating the need to store and compute masked attention weights and key/query scores. It also optimizes training by reducing the number of activations recalculated during the backward pass. LLaMA-2~\citep{llama2} focuses on fine-tuning a safer and more effective LLaMA-2-Chat model specifically for dialogue generation. The pretrained model leverages 40\% more training data with a larger context length and employs grouped-query attention for improved performance. More recently, LLaMA-3~\citep{llama3} released a family of LLMs, ranging in size from 8 to 70 billion parameters. These models come pretrained and instruction-tuned variants. The instruction-tuned variant  outperform a number of open-source chat models on the industry benchmarks.

\begin{table}[!tbp]
  \centering
  \begin{tabularx}{\textwidth}{l|l|l|l|p{2cm}|X}
    \hline
     \multicolumn{1}{c|}{Type}& \multicolumn{1}{c|}{Name} & \multicolumn{1}{c|}{Release} & \multicolumn{1}{c|}{\#Parameters} & \multicolumn{1}{c|}{\#tokens} & \multicolumn{1}{l}{Training Data} \\ \hline
     \multirow{6}{*}{\STAB{\rotatebox[origin=c]{90}{Encoder}}}
      & BERT~\citep{BERT-19}     &   2018     & 110, 340M   & 137B & en-Wiki, Books\\
      & RoBERTa~\citep{RoBERTa}  &   2019     & 355M        & 2.2T & en-Wiki, Books, Reddit, CC-news \& stories\\
      & ALBERT~\citep{albert-20} &   2019     & 12,18,60M   & 137B & en-Wiki, Books \\
      & ELECTRA-base~\citep{ELECTRA}  &  2020      & 110M  &      & Wikipedia and Books  \\
      & DistilBERT~\citep{DistilBERT}  &  2020      & 66M       &  137B     & en-Wiki and Toronto Book Corpus\\
      & DeBERTa~\citep{Deberta}  &  2021      & 1.5B        &      & en-Wiki, Books, CC-stories, Reddit\\
      \hline 
      %& XLNet        &                    &                   &           &           \\
      \multirow{26}{*}{\STAB{\rotatebox[origin=c]{90}{Decoder}}}
      & GPT-1~\citep{gpt1}     &  2018     & 117M          & 1.3B      & Books corpus \\
      & GPT-2~\citep{gpt2}     &  2019     & 1.5B          & 10B       & Reddit       \\
      & GPT-3~\citep{GPT32020} &  2020     & 175B          & 300B      & Wikipedia, CC 2016-19, Books, WebText\\
      & GPT-NeoX~\citep{GPT-NeoX} &  2022     &     20B      &   472B   & PILE and C4\\
      &Jurassic-1~\citep{lieber2021jurassic} &  2021     &    7,178B   &     & PILE \\
       &  Gopher~\citep{Gopher}  & 2021         & 280B           & 300B      & C4, books, news data, MassiveWeb, GitHub, en-Wiki\\
        &  LaMDA~\citep{LaMDA22}  & 2022         &  2,8,137B      &  1.56T       & Dialogues data, subsets of C4, en\& non en-Wiki, programming QA\\
        &  BLOOM~\citep{BLOOM-22}  & 2022         & 176B     &  366B       & ROOTS datasets\\
        & PaLM~\citep{PaLM}       &  2022     &   8,62,540B    & 780B     & Web \& conversation data, books, Wiki, GitHub code\\
        &  Grok-1~\citep{grok1}  & 2022     & 176B     &  366B       & ROOTS datasets\\
        &  ChinChilla~\citep{chinchilla22}  &  2022  &  137B  &   1.4T & MassiveText\\
        &  M-T NLG~\citep{megatron-turing-nlg} &  2022       &  530B &  3.5T      & Books3, OpenWebText2, Stack Exchange, GitHub, PubMed Abstracts, Wikipedia, Gutenberg, BookCorpus2, RealNews NIH ExPorter,ArXiv, Pile-CC, Stories-CC\\
        &  Falcon~\citep{Falcon} & 2023        &  7,40,180B &  3.5T      & RefinedWeb\\
       &  PaLM2~\citep{palm2}  & 2023         & 340B           & 3.6T      & Web data, mathematics, books, Wiki, GitHub code,  \\
       &  &&&   & conversation data \\
      & GPT-4~\citep{GPT4}     &  2023     & 1.76T         & 13T      &        \\
      & Mistral~\citep{mistral7b}& 2023   &  7.3B          &           &           \\
      & LLaMA~\citep{llama23}   &  2023    &  7,13,33,65B  & 1-1.4T   & Web data\\
      & LLaMA2~\citep{llama2}   &  2023   &  7, 13,34,70B  & 2T       & Web data \\
      & LLaMA3~\footnote{\url{https://huggingface.co/blog/llama3}} & 2024  & 8B, 70B      & 15T       & Web data\\
      & Phi3~\citep{phi3} & 2024  &  3.8B   & 3.3T   & Filtered web\& synthetic data\\ 
      & YOCO ~\citep{Yoco} & 2024  &  13B   &  1.6T  & \\ 
      & Falcon2 ~\citep{falcon2} & 2024  &  11B  &  5.5T  & RefinedWeb \\ 
      \hline
      \multirow{8}{*}{\STAB{\rotatebox[origin=c]{90}{Encoder-Decoder}}}
      & MASS~\citep{mass}    & 2019 &   &   &  News Crawl Datasets   \\
      & T5-base~\citep{T5-p}      & 2019 &  223M    & 156B & CommonCrawl\\
       & BART-base~\citep{BART22}  & 2019  &  139M &  & BookCorpus, Wikipedia \\
      & mT5-base~\citep{mt5-mc4}  & 2020 & 300M    &  & CommonCrawl 101 languages\\
      & CPM-2~\citep{CPM-2}  & 2021 & 11B    &  & WuDaoCorpus \\
      & UL2~\citep{flan-UL2} &   2022 &   20B &  1T     &  C4  \\
      & AlexaTM~\citep{AlexaTM} & 2022   &   20B     &  1.319T   & Wikipedia and mC4 \\
       & Gemini~\citep{Gemini}&   2023 & 1.8, 3.25B   &     & Books, Web documents, code, audio-video data \\
      \hline
      \multirow{8}{*}{\STAB{\rotatebox[origin=c]{90}{Sparse Models}}}
      & CPM2-MoE~\citep{CPM-2}  & 2021  &  198B &   &  WuDaoCorpus  \\
     & GLaM~\citep{GLaM}   & 2022  &  &   &    \\
      & Switch-C~\citep{Switch}   & 2022  & 1.5T & 500B  &  mC4  \\
      & GShard-M4~\citep{Gshard}   & 2020  & 600B &   &  Web documents for 100 languages   \\
      & Mixtral~\citep{MoE}   & 2023 & 46.7   &   & Instruct Dataset\\
      & GLaM~\citep{GLaM}   & 2022  & 1.2T  & 1.6T  & Filtered Webpages, Wikipedia, news, and books   \\
      & DeepSeekMoE~\citep{deepseekmoe}   & 2024 & 67B   &   & Chinese \& English  Web text, Math, code, scientific literature etc.   \\
      & DeepSeekMoE-V2~\citep{deepseekv2}   & 2024 & 236B  & 8.1T   & Same as DeepSeekMoE   \\
      \hline
  \end{tabularx}
  \caption{An overview of Large Language Models (LLMs) is presented. Here, M, B, and T denote million, billion, and trillion, respectively. While most LLMs are multilingual, their primary focus is often the English language. "en-Wiki" denotes an English Wikipedia dump. This work focuses on general-purpose LLMs; for a domain-specific survey on LLMs, multiple domains can be referred to in~\citep{llm-domains-survey}.}
    \label{tab:overview_llms_relatedwork}
\end{table} 

YOCO~\citep{Yoco} is released more recently, which is a decoder-decoder model that only caches key-value pairs once. It consists of cross-decoder stacked upon a self-decoder. Although, the model is similar to decoder-only architectures but main difference is it only caches once which reduces GPU demands  substantially. YOCO yield comparable performance to SOTA LLMs in different settings such as scaling of model size and training tokens. Another interesting contribution is the context length of of YOCO which is extended upto 1 million. 

Falcon2 ~\citep{falcon2} is recently released decoder only model, trained on RefinedWeb dataset using 5.5 trillion tokens with context length of 8K. Except English, it support German, French, Spanish, Italian, Swedish, Dutch, Portuguese, Czech, Polish, and Romanian languages. It outperforms Llama3 8B on two NLU benchmarks.

\subsection{Encoder-Decoder Models}
Encoder-decoder models are suitable for tasks requiring understanding and responding to input text. A notable example is T5~\citep{T5-p}, which employs a unified text-to-text training approach for various NLP tasks. It leverages a unique type of MLM where entire spans of consecutive words are masked, accelerating training by dealing with shorter sequences. T5 also utilizes adapter modules to adjust to specific NLP tasks during fine-tuning.

mT5~\citep{mt5-mc4} is a multilingual LLM built upon T5. It aims to solve the issue of "accidental translation" in zero-shot setup, where the model wrongly translates parts of the output into the wrong language. It utilizes a mixing approach that incorporates unlabeled pretraining data during fine-tuning. This process effectively reduces the occurrence of accidental translations. To handle multiple languages, mT5 employs a large vocabulary size. Additionally, it utilizes data sampling techniques to avoid underfitting or overfitting for specific languages.

UL2~\citep{flan-UL2} implements a Mixture-of-Denoisers (MoD) objective within an encoder-decoder architecture. This approach allows for more focused optimization by introducing the "mode switching" technique. This technique enables tailoring the fine-tuning of the model for a downstream task to a particular pretraining scheme within MoD. UL2 presents a powerful and versatile framework for pre-training LLMs. It offers superior performance across various NLP tasks and demonstrates potential for further advancements in LLM capabilities. The model surpasses other models, including GPT-3~\citep{GPT32020} and T5~\citep{mt5-mc4} using zero-shot and one-shot learning on a wide range of NLP tasks. When scaled to 20 billion parameters, it achieves top performance on 50 benchmark NLP tasks. 

AlexaTM~\citep{AlexaTM} is a powerful multilingual sequence-to-sequence model20-billion-parameter LLM with impressive capabilities. Multilingual Expertise: It can understand and process information across 100 languages. It surpasses GPT-3 and other larger models on benchmark tasks such as QA, summarization, and translation, especially for low-resourced languages by taking the advantage 1-shhot or few-shot learning. The success of AlexaTM is attributed to a combination of causal language modeling and denoising tasks during pretraining.

BART~\citep{BART22} utilizes a denoising autoencoder approach with a standard sequence-to-sequence translation model architecture. It is pretrained by corrupting text with an arbitrary noise function and then learning to reconstruct the original text. BART yield superior performance in text generation and comprehension tasks. The study explores various text corruption methods, finding the best performance with a combination of random sentence shuffling and a novel "in-filling" scheme, where text sections are replaced with a mask token. It achieves performance comparable to RoBERTa on several benchmarks using similar training resources. Additionally, it surpasses existing models on QA, summarization, and abstractive dialogue tasks. 

Gemini~\citep{Gemini} introduces a new family of multimodal models exhibiting promising capabilities across various domains, including image, audio, video, and text understanding. The Gemini architecture is built on top of Transformer decoders and is trained to support a 32k context length through efficient attention mechanisms.

\subsection{MoE Models} 
Unlinke dense models, MoE models provide a sparse architecture for effective training of big models. The MoE based models achieve performance that is comparable to dense models and permits large increases in model size without requiring a large amount of processing power. This is a result of MoE only ever activating a certain subset of experts at once. Such models mainly follow decoder-only~\citep{MoE,GLaM} and encoder-decoder~\citep{Switch,Gshard} architectures. 

CPM-2~\citep{CPM-2} is cost-efficient model, explores pretraining bilingual models (English and Chinese) using a MoE architecture. It also introduces a memory-efficient framework called INFMOE for large-scale inference. CPM-2 presents two LLMs based on MoE and encoder-decoder architectures, respectively. The CPM-2 encoder-decoder model is bilingual (English-Chinese) with 11 billion parameters, while the MoE-based model is larger, with 198 billion parameters. To accelerate training for both models, the researchers split the process into three stages. Notably, CPM-2 outperforms mT5 on several NLP tasks.

Switch~\citep{Switch} Transformer based on MoE encoder-decoder architecture reduces cost and training time by using a simpler approach for handling several model components during training. These improvements, which build upon existing models, enable up to 7x faster pretraining without the need for additional resources. The advantages spread to multilingual setup, where all 101 tested languages perform better. Switch is pretrained successfully with trillion-parameters on a large dataset, pushing the limits of LLM size and achieving a 4x speedup over baseline models. This work provides a significant improvement in the stability, scalability, and efficiency of LLM training.

GShard~\citep{Gshard} leverages a MoE approach and provides a module that simplifies the parallel processing for training LLMs with the ability to handle upto 600 billion parameters. This module divides the training computations automatically by eliminating the need for manual configuration. GShard presents an interface that requires minimal changes to the existing model code for scaling.

Mixtral-8x7B~\citep{MoE} leverages a MoE approach with eight experts per layer. This allows each token to access billions of parameters while actively using only a fraction (i.e., 13 billion) during processing. The Mixtral-base model yields strong performance, matching or exceeding theleading models like GPT-3.5 and Llama-2 on various tasks. It particularly excels in areas like code generation, math, and handling multiple languages. Mixtral-8x7B-Instruct shines in following instructions, surpassing other chat-focused models.

Generalist Language Model(GLaM)~\citep{GLaM} family utilizes a sparsely activated decoder-only MoE architecture, significantly reducing the training cost compared to dense models by activating only a subset of experts for each input token. Its MoE based architecture scales the model capacity while incurring  less training cost to dense models. The largest model has 1.2T parameters, which is approximately 7X larger than GPT-3. However, it consumes 1/3 of the energy used as compare to GPT-3 and requires half of the computation flops for inference. the GLaM yield better  zero-shot and one-shot performance in a number of NLP tasks.

Recently, DeepSeekMoE~\citep{deepseekmoe} designed for highly specialized LLM, using two main strategies: 1) dividing experts into smaller groups and activating a chosen subset for more flexible combinations and 2) isolating shared knowledge experts to reduce redundancy. It demonstrates good performance across various scales, even with fewer parameters and computations compared to existing models GShard~\citep{Gshard}, LLaMA2~\citep{llama2}. DeepSeekMoE achieves comparable or superior performance, highlighting its potential for building powerful and efficient LLMs. 

More recently released DeepSeek-V2~\citep{deepseekv2}  is a game-changer for LLMs, achieving high performance while being economical to train and use. It utilizes MoE approach and yet powerful despite fewer parameters, along with techniques that significantly reduce memory usage and boost generation speed. This efficient model surpasses its predecessor in all aspects.

%Few EU languages including Croatian, Estonian, Irish, Latvian, Lithuanian, Maltese, Slovak, and Slovenian stand among the less-resourced languages and there is not any pretrained LLM. 

%Monolingual datasets for training European language models 
\section{Pretraining Datasets}
\label{pre_datasets}

A massive amount of high-quality data is a core ingredient for training foundation models. This section presents an overview of existing resources for training general-purpose LLMs for official EU languages, along with statistics and their main sources of creation. We present subsets of multilingual datasets along with statistics for specific language corpora. Table~\ref{tab:pretraining_mono_datasets} and Table~\ref{tab:multilingual_datasets} present overviews of monolingual and multilingual pre-trained datasets, respectively. For English, a recent survey~\citep{datasets-survey-llms} provides a comprehensive overview of available datasets.
\subsection{An Overview of European Languages}
The EU is home to a diverse array of languages and cultures. A vital component of the EU's cultural legacy is its linguistic diversity, which is actively encouraged by the organization through its institutions and activities. There are 24 official languages\footnote{\url{https://european-union.europa.eu/principles-countries-history/languages_en}} in the EU, but only three—English, French, and German—are recognized  by the European Commission as procedural languages. 
\begin{figure}[!h]
    \centering
    \includegraphics[width=\textwidth]{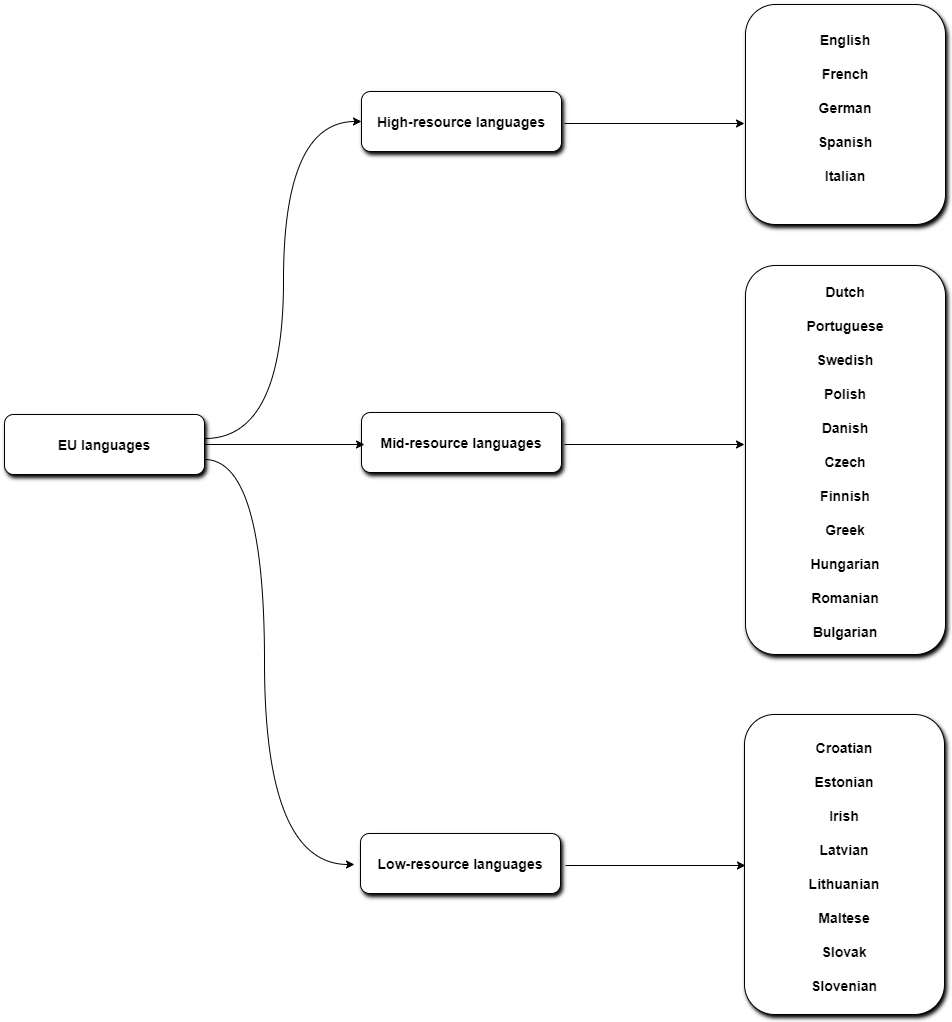}
    \caption{Categorization of EU languages in low-resource, mid-resource, and high-resource languages}
    \label{fig:eu-languages}
\end{figure}
In contrast, the European Parliament recognizes all official languages as working languages. In terms of language resources, EU languages can be broadly categorized into low-resource, mid-resource, and high-resource categories.  This classification is based on the availability of language resources, such as large unlabelled corpora, annotated datasets, and  linguistic tools for NLP tasks. 
\begin{table}[!h] 
    \begin{tabularx}{\textwidth}{|l|l|l|l|X|p{2cm}|}
    \hline
         Language  & Size  & Tokens  & LLM  & Source\\
         \hline
          German  &  163.4GB &   & GBERT\cite{German-llm}  &  de-OSCAR, Wiki\\
          &   &   & GELECTRA  & de-OPUS, legal text \\
          &  145GB &   & GottBERT\cite{GottBERT-llm}  &  de-OSCAR, Multilingual corpora \\
          &   & 65B  & LeoLM~\citep{leollm} & de-OSCAR-2301\\
          \hline
          French  & 138GB  & 32.7B  &  CamemBERT\cite{CamemBERT-french}  & fr-OSCAR\\
          & 135GB  &  31.9B  &    & fr-CCNet \\
          & 4GB  & 990M  &    & fr-Wikipedia\\
          & 71GB  &  &  FlauBERT\citep{flaubert-french}  & Web, Wiki, books \\
          & 135GB  &  32B  & PAGnol-XL\citep{CroissantLLM-french}    &  fr-CCNet \\
          & 1.1TB (UC)  & 78.7B  &   Cedille\citep{Cedille-french}  & fr-mC4 \\
           &   & 6.97T  &    & French-PD-Newspapers \\
          
         \hline
          Italian  &  13GB &   & GePpeTto\citep{GePpeTto-italian}  &  ItWac corpus, it-Wiki \\
          & 215GB   & 40B  & IT5\citep{IT5-italian}  &  it-mC4 \\
          &   &   & Fauno\citep{Fauno-italian}    &  Synthetic data, translated Quora etc.\\
          &   &   & Camoscio\citep{Camoscio-italian}    & finetuned LLaMA-7B   \\
          &  215GB  &    &  BART-IT\citep{BART-italian}  & Italian mC4  \\
          &  135GB  &    &  LLaMAntino\citep{LLaMAntino-italian}  & it-OSCAR   \\
         \hline
          Spanish  &  570GB  &    &  MarIA\citep{MarIA-spanish-llm-23}  & National Library Spain 2k9-2k19  \\
          &  &  & Spanish-BERT\cite{BERT-spanish-23}  & OPUS and Wikipedia\\
         \hline
          Polish  &   & 1.8B  & pl-Corpus~\citep{polish-corpus-11}  &  NKJP-Journalism, books, social blogs etc.\\
          & 135GB  &   &  BERT~\citep{Transformer-Polish}  & pl-CC, Wiki, OPUS, CLARIN etc.,\\
          &   & 8.599B  & HerBERT~\citep{HerBERT-Polish}   & NKJP, Wiki, CCNet, OpenSubtitles\\
          \hline
          Romanian & 12.6GB  & 2.07B  & RoBERT~\citep{Romanian-BERT-20}   & Romanian Wiki, RoTex, OSCAR  \\
          & 12.02GB  & 2.56B  & RoGPT-2~\citep{Romanian-GPT2-21}   & ro-Wiki, OSCAR, books, news \\
          & 15.4GB  &   &  ALR-BERT~\citep{lite-romanian-bert-22} & OPUS, OSCAR, Wiki  \\ 
          \hline
          Dutch  & 39GB  &  6.6B &  RobBERTa~\citep{Dutch-RoBERTa}  & nl-OSCAR  \\ 
          &   &  33B &  GPT-NEO~\citep{gptneo-dutch}  & nl-mC4 \\
         & 234GB  &  40B &  & nl-mC4~\citep{dutch-lrs-23}, NL-CC, Forums, books~\citep{dutch-Giga} \\
        \hline  
         Greek  & 29.21  & 3.04B & Greek-BERT\citep{GREEK-BERT20}    & el-EuroParl, Wiki, OSCAR \\        
         & 76.9GB   & & GreekBART\citep{GreekBART23}     & el-EuroParl, CC, OSCAR  \\
         & 50GB & 3B  & GreekT5, GreekBART   &   el-Web Corpus~\citep{lampos2004archiving} \\
        \hline  
        Hungarian &  & 1.5B, 9B  &   &   hu Gigaword~\citep{Hungarian-giga-word}, Webcorpus2.0 ~\citep{hungarian-corpus} \\
        & 314GB &  41.50B & GPT-3~\citep{Hungarian-gpt23}   & hu-CC, Wiki, OpenSubtitles \\
        \hline  
        Swedish & 17.9GB  & 3.497B  & KB-BERT~\citep{swedish-bert20} &  sv-Wiki, legal, news, social media \\
        &  100GB &    & GPT-SW3~\citep{GPT-swedish22} & sv-Web-data, OSCAR  \\
        &   &    &  & Flashback, Wiki, Subtitles   \\
        &   & 5.743B   &  SweCTRL-Mini~\citep{SwedishCTRL-mini23} & sc-mC4, and Project Runeberg  \\
         \hline 
         Bulgarian & 50GB  &   & GPT\&BERT~\citep{GPT-WEB-BG23} & Bulgarian Web corpus \\
        \hline 
         Czech & 72GB  &  10B & Czech-ALBERT~\citep{Albert-Czech-2020} &  csTenTenl7(Web-crawl 2k15-2k17\\
         & 36.9GB  &   & Czert-BERT~\citep{BERT-Czech-2021} &  cs-wiki,  National corpus~\citep{syn-v4-czech}, News \\
         &   &  4.917B & RobeCzech~\citep{RobeCzech-21}  & cs-wiki,SYN-v4, Czes, W2C \\
        \hline
         Portuguese  &   &  2.7B  &  & Web Corpus~\citep{brWaC-corpus}\\
         &   & 7.8B  & Sabia \&LLaMA 7B,65B~\citep{Portuguese-llm-23} & pt-Corpus ClueWeb-2022~\citep{Clueweb22}\\
        \hline
         Danish &   & 1B  &  & Danish Gigaword Corpus\\
        \hline
        Finnish  & 84GB  & 1B  & GPT~\citep{gpt-finnish-22} & Yle, Kielipankki, Web crawl\\
        & 84GB  & 0.78B  & GPT-2~\citep{gpt2-finnish}  & fi-Wiki, News Archive, Suomi24 \\
        &   &   & BERT~\citep{bert-finnish-19}  & Suomi24\\
        &   & 38B   & FinGPT~\citep{FinGPT}  & fi-CC, Yle, Wiki, mC4, ePub, Lehdet, Suomi24, Reddit, STT, ROOTS etc.,  \\
        \hline   
        Slovak  & 17.4GB  &   & SlovakBERT\citep{SlovakBERT22}  & sk-Wiki, OpenSubtitles, OSCAR  \\
        %Lithuanian  & 14.5GB  & 1.67B  &  & OSCAR v23.1 \\
        \hline
        Slovene  &   & 4.20B  & SloT5~\citep{Slovenian-T5-23}  & Gigafida, Janes, KAS, SiParl, SlWaC \\
          &   & 1.33B  & SloBERTa~\citep{SloBERTa-21}   & Gigafida 2.0, Janes \\
          \hline
          Estonian&   &   & GPT2 \& EstBERT~\citep{EstBERT-estonian}   & Estonian National Corpus \\
          \hline
          Maltese &   & 46.66B  & BERT~\citep{BERT-Maltese-22}   & Korpus Malti V4.0 \\
    \hline
    \end{tabularx}
    \caption{Pretraining monolingual datasets for EU languages. Few recent survey papers can be referred for the English\citep{datasets-survey-llms}. Language code with dataset denote the its subset, Wiki denotes the Wikipedia dumps}.
    \label{tab:pretraining_mono_datasets}
\end{table}
%Split the table monolingual  and multilingual  after collection of data

\subsection{Monolingual pretraining datasets}\label{subsec:mono-pre_data}
Transformer models require massive amounts of data to develop robust LLMs. This necessitates the use of large, high-quality datasets for pretraining, along with sophisticated finetuning datasets and language understanding evaluation benchmarks. This section provides an overview of publicly available monolingual pretraining datasets for EU languages. Table~\ref{tab:pretraining_mono_datasets} shows the details of these datasets. 
%OSCAR is a set of monolingual corpora extracted from Common Crawl. 
\textbf{German:} %
German is considered a high-resource language, having a substantial amount of LRs for NLP tasks.  The German deWaC corpus is a subset of the WaCky~\citep{WaCky} corpus, crawled from the web. It contains approximately 1.7 billion tokens and includes multi-domain text, such as news, blogs, forums, and others. Several publicly available datasets offer valuable resources for training German language models. The OSCAR~\citep{oscar} corpus is a large multilingual text dataset extracted from Common Crawl (CC). The German subset of OSCAR is particularly noteworthy, measuring 496.7GB in size and containing approximately 7 billion documents and nearly 4.68 trillion words. Open Legal Data~\citep{German-legal} is a specific dataset catering to the legal domain. It offers approximately 2.4GB of German court decisions. Moreover, other multilingual corpora like Wikipedia~\citep{wikidump} dumps, CC~\citep{CCNet}, and OPUS~\citep{OPUS} all contain German subsets. These subsets provide a vast amount of text data in various domains.

\textbf{French:} %
French stands among the high-resource EU languages, boasting a substantial amount of LRs, for wide range of NLP tasks. More recently, the French-PD-Newspapers dataset\footnote{\url{https://huggingface.co/datasets/PleIAs/French-PD-Newspapers} } is released as a valuable addition, containing three million newspapers and periodical editions digitized by the French National Library. This massive collection offers a variety of French text, totaling an impressive 6.97 trillion words. While other resources further enrich the French NLP landscape. The French subset of CulturaX~\citep{culturax} offers 319.33 billion tokens, while OSCAR's French subset provides 382.2 GB of data, containing approximately 4.17 trillion words. Moreover, large portion of unlabelled French text in multilingual datasets like CCnet, Wikipedia, and mC4 offer valuable resources. 

\textbf{Italian:} %
Italian stands out as a high-resource language with substantial amount of LRs available for NLP tasks. One prominent monolingual resource is the WaCky corpora~\citep{WaCky}. This web-derived collection, compiled between 2005 and 2007, contains an Italian subset (itWaC) exceeding 1 billion words. Looking beyond WaCky, the Italian part of OSCAR consists of 229.3 GB, containing 28 million documents and 2.42 trillion words. Moreover, the Italian subsets within Wikipedia and mC4 are also noteworthy resources for building Italian LLMs. 

\textbf{Spanish:} %
MarIA~\citep{MarIA-spanish-llm-23}, a family of Spanish LLMs leverages a massive dataset for training. This dataset includes 570GB of clean and deduplicated text with 135 billion words extracted from the Spanish Web Archive crawled by the National Library of Spain between 2009 and 2019. Other resources like the Spanish subset of OSCAR, containing approximately 5.13 million documents and 4.28 trillion words. Similarly, the Spanish parts of Wikipedia dumps and OPUS offer substantial amounts of text data.

\textbf{Polish:} %
The National Corpus of Polish (NKJP) ~\citep{polish-corpus-11} presents a large collection of texts of various sizes in the main corpora. The sources of these corpora include newspapers, classic literature, journals, specialist periodicals, transcripts of conversations, and web texts. The NKJP corpus consists of 1.5 billion words and about 1.8 billion segments, with each segment being a technical linguistic term; for example, words are considered single segments. The Polish subset in OSCAR consists of 1.93 billion documents and 1.25 trillion words. Web text and CC~\citep{Transformer-Polish} were crawled and filtered, resulting in a size of 135GB. This includes subsets of the web corpus, Common Crawl from 2019 to 2020, publicly available Polish text, Wikipedia, the Parliamentary Corpus, and a few other smaller corpora obtained from books, CLARIN, OPUS, and articles. Moreover, Open Subtitles also contains a Polish subset.

\textbf{Romanian:} %
RoText~\citep{rotext} is a monolingual, quality-filtered corpus of Romanian language texts crawled from various web resources. For quality filtration, page numbers of PDF files and duplicate text from headers and footers are removed. RoText consists of 240 million words. Additionally, subsets of Romanian text from multilingual corpora including OPUS, OSCAR, and Wikipedia dumps consist of 635 million, 1.78 billion, and 60.5 million words, respectively.

\textbf{Dutch:} %
Gigacorpus\footnote{\url{http://gigacorpus.nl/}}, a freely available corpus consisting of approximately 40 billion tokens, constitutes the largest monolingual Dutch corpus for training LLMs. It is comprised of a combination of books, historical and fictional novels, the Dutch news corpus, Web-based news articles, the multi-genre SoNaR-500 reference corpus, contributing a total of 2.4 billion tokens. Moreover, Dutch subsets in OSCAR contain 1.23 trillion tokens. Additionally, CC and Wikipedia dumps also include a portion of Dutch text.

\textbf{Greek:} 
The main Greek textual resources include the Greek part of OSCAR, Wikipedia, Europarl~\citep{europarl}, and a clean version of Common Crawl~\citep{GREEK-BERT20,GreekT5-23,GreekBART23}. The latest version of OSCAR consists of 503.12 billion tokens of Greek language.

\textbf{Hungarian:} 
There are several large collections of Hungarian text datasets\footnote{\url{https://github.com/oroszgy/awesome-hungarian-nlp?tab=readme-ov-file##raw-corpora}}. The datasets include Hungarian Webcorpus, containing 1.48 billion words in an unfiltered format, and is freely available. An even larger option is Hungarian Webcorpus 2.0, which contains over 9 billion words derived from web data. OSCAR is a massive collection of text in many languages, including Hungarian, with 2.34 billion words. For Hungarian-specific purposes, there's emLam~\cite{hungarian-emLam17}. Moreover, there are general purpose corpora that include Hungarian text, such as Leipzig corpora~\citep{leipzigcorp}, Hungarian part of CC, Wikipedia, and OpenSubtitles.

\textbf{Swedish:} %
Swedish stands among the high-resource EU languages due to the substantial amount of unlabeled and other language resources available. Project Runeberg~\citep{SwedishCTRL-mini23} is a valuable digital archive dedicated to providing open access to Nordic literature. It focuses on making cultural and historical texts in Swedish and other Nordic languages readily available online, particularly those that are difficult to obtain elsewhere. In addition to Project Runeberg, several other resources offer valuable datasets, including the OSCAR Swedish Subset, consisting of approximately 7.54 million documents and 507 billion tokens, and subsets within mC4, PILE, and Wikipedia dumps offer additional large corpora for LLM training.

\textbf{Bulgarian:}  %
Two publicly available datasets provide valuable resources for training LLMs for Bulgarian. Bulgarian Web Dataset~\citep{GPT-WEB-BG23} comprises over 50GB of cleaned and balanced online textual data published from 2015 to 2021. The data covers various domains, including books, social media, and scientific literature. Bulgarian Subset of OSCAR Corpus contains 35.1GB of text data. This subset includes approximately 2.88 million documents and 240 billion words. Both datasets can be used for pretraining LLMs for Bulgarian, either independently or in conjunction with other textual resources like Wikipedia.

\textbf{Czech:} %
Several publicly available corpora provide resources for training Czech language models. SYN-V4 Corpus~\citep{syn-v4-czech} is a traditional corpus, boasts rich bibliographic metadata alongside nearly 4.3 billion tokens of text. Primarily focusing on newspapers, it also includes fiction and non-fiction content, covering the period from 1990 to 2014. CsNews~\citep{BERT-Czech-2021} corpus is also a collection of crawled Czech news articles, reaching a total size of 7.8GB. Moreover, Czes~\citep{czes11} consists of 432 million tokens of text from newspapers and magazine articles.
csTenTen17~\citep{csTenTen17} corpus Compiled from downloaded texts between 2015 and 2017, this large corpus consists of 10.5 billion words. Notably, this is double the size of its predecessor from 2012.

\textbf{Portuguese:} %
Brazilian Portuguese is considered a low-resource language~\citep{portuguese} when it comes to language resources. A web corpus (brWaC)~\citep{brWaC-corpus} is developed by filtering a large collection of crawled webpages. The filtering process removes duplicates and ensures the inclusion of web-pages from diverse domains. The brWaC corpus is a valuable resource for researchers, consisting of 145 million sentences and 2.7 billion tokens.

\textbf{Danish:} The Danish Gigaword Corpus~\citep{danish-gigaword} is a diverse and publicly available corpus of one billion words. It covers a wide array of domains, including legal texts, social media, conversations, web content, Wikipedia entries, books, news articles, and more.
%%%\textbf{Bulgarian:} no resoures for llms

\textbf{Finnish:} %
The Finnish corpora cover various sources, including news articles, books, social media, and crawled web data. Yle News Corpus~\citep{FinGPT} is widely used dataset consists of nearly 800K articles, 220 million tokens crawled from the Yle News website, covering both international and local news. The Suomen Tietotoimisto (STT) Dataset~\citep{FinGPT} contains 2.8 million news articles (around 300 million tokens) collected by the Finnish News Agency from 1992 to 2018.

Social Media datasets include Suomi24, which used to train FinGPT is comprised of nearly 95 million comments and 5 billion words curated from the Finnish social networking site Suomi24 between 2001 and 2020. The dataset is available on the Language Bank of Finland.
Reddit-Fi Corpus~\citep{FinGPT} is a collection of Finnish text from Reddit posts contains nearly 4 million comments and 150 million tokens, spanning from 2009 to 2022.

The large multilingual dataset with Finnish Subset include mC4 ~\citep{mt5-mc4}. This subset contains 8 billion tokens across 19 million documents in the Finnish language. The CC Finnish Crawl~\citep{FinGPT} is  custom extraction from Common Crawl covers Finnish language text from 2013 to 2022, resulting in a large collection of 20 billion tokens.
ROOTS Finnish Subset~\citep{Roots} represents approximately 0.03\% of the overall ROOTS dataset. Moreover, Fi-Wiki dataset consists of 110 million tokens extracted from the Finnish Wikipedia dump. Parsebank~\citep{finnish-parsebank} is a large corpus for the Finnish language, collected from CC during 2015-2016. The dataset is cleaned, deduplicated at the paragraph level, and consists of 6 billion tokens.

An ePub corpora\footnote{\url{https://kansalliskirjasto.finna.fi/}} is a collection of online published books maintained by the National Library of Finland. The ePub corpus is a collection of nearly 30K electronic books in the Finnish language. Another dataset, Lehdet~\citep{FinGPT}, is also based on archived HTML material collected by the National Library of Finland and includes news crawls. However, due to copyright restrictions, both datasets are not publicly available. Lönnrot~\citep{FinGPT} dataset is created by collecting Finnish and Swedish out-of-copyright literature, which contain a total of 125 million tokens. The Yle news corpus~\citep{FinGPT} is crawled from the Yle News website, which features both international and local news. This widely used dataset is employed to train Finnish LLMs, including FinGPT~\citep{FinGPT, gpt-finnish-22}. It consists of nearly 800K articles, resulting in 220 million tokens. Suomen Tietotoimisto (STT) dataset is also a Finnish News Agency~\citep{FinGPT}. The collected datasets from the 1992-2018  contains 2.8 million news articles which resulted around 300 million tokens. Moreover, Lönnrot Dataset~\citep{FinGPT} contains 125 million tokens of Finnish and Swedish out-of-copyright literature.

\textbf{Slovak:} %
Being a low-resource language, Slovak has limited publicly available resources. One exception is the web corpus used to train the SlovaKBERT model~\citep{SlovakBERT22}. This corpus leverages Wikipedia, Open Subtitles, and OSCAR for data. Additionally, the web corpus was built by applying language detection to extract clean text, including titles and main content, from each webpage. Combining the available and crawled corpora (sk-Wiki, Open Subtitles, and OSCAR) yields a total of 17.4GB of cleaned text.

%%%\textbf{Lithuanian}
\textbf{Estonian:} The Estonian National Corpus~\citep{EstonianNC} is a collection of texts from various sources. It mainly consists of four subsets i). Estonian Reference Corpus is based on newspaper articles, fiction, science, and legislation texts makeup about 242 million words. ii).  Estonian Web Corpus is the  text downloaded from the internet. iii). Estonian Wikipedia Corpus is a snapshot of the Wikipedia dumps, containing around 38 million words. Moreover, OSCAR consists of around 80 billion words. 

\textbf{Maltese:}  Korpus Malti~\citep{BERT-Maltese-22} is curated from specific sources instead of scraped randomly from the web resulted 46.66 billion tokens\footnote{\url{https://huggingface.co/datasets/MLRS/korpus_malti}}. The web sources include literary works, official newsletter, non-fictional texts published by the University of Malta. This approach creates a smaller but higher quality dataset compared to web scraping. 

\textbf{Slovene:} %
Few publicly available datasets for training Slovenian language models cover various domains, including fiction, textbooks, news, academic writing, and parliamentary debates.

Gigafida 2.0 Corpus~\citep{Gigafida-slovene-corp} is a general general cCorpora, builds upon previous versions (including Gigafida 1.0) to create a large collection of Slovenian text. The dataset focuses on standard language, excluding non-standard variations. It is filtered, deduplicated, and includes fiction books, textbooks, and news articles. SlWaC Corpus~\citep{slWac-corpus} crawled from the web is a good source of unlabeled text. It consists of 92 million documents, resulting in 38 billion words.

KAS Dataset~\citep{KAS2.0-slovene-corp22} is a collection of Slovenian academic writing from 2000 to 2018. It includes nearly 82,000 individual theses, resulting in 1.5 billion tokens. The theses are gathered from digital libraries of Slovenian higher education institutions. siParl Corpus~\citep{siparl} is the collection of Slovenian parliamentary debates from 1990 to 2018. It includes over 8K sessions, one million speeches, and a staggering 200 million words. The dataset provides rich metadata, including details about speakers, session types, and even structural annotations. 

\subsection{Multilingual  Pretraining Datasets}\label{subsec:multilingual_data}
This section provides an overview of publicly available multilingual text datasets for training foundation LLMs. Large subsets of the corpus contains European languages. Table \ref{subsec:monolingual_data} show the detail of these datasets. 

\textbf{OSCAR:}~\citep{oscar} Short for \textit{Open Super-large Crawled Aggregated Corpus} is a large multilingual dataset built by processing the Common Crawl. The size varies depending on the version. OSCAR provides the data in both original and deduplicated formats. The original format includes all the data after language separation. The deduplicated format removes redundancies within the language-specific data. OSCAR 23.01 is the latest version, offering large amounts of unannotated raw web data for pre-training LLMs. It emphasizes data quality for web-based corpora and includes data for low-resource languages.

\textbf{Wikipedia:} It is a collection of cleansed articles from Wikipedia in all languages. The Wikipedia dump is used to create the datasets, with one split for each language. Every example includes the whole content of a single Wikipedia article that has been cleaned to remove unnecessary sections and markdown.

\textbf{CCNet:} FastText plays a crucial role in identifying the languages within the data, while a hashing technique ensures duplicates are eliminated. This meticulous process allows CCNet to focus on delivering high-quality, monolingual data

\textbf{mC4:} A multilingual variant of the C4~\citep{C4} dataset called mC4 ~\citep{mt5-mc4}, comprises text data in 101 languages scraped from Common Crawl. It is a massive collection of text in over 100 languages, derived from the popular CC web crawl corpus.  The size of mC4 reaches is 38.49 TB. Unlike the CC dataset, mC4 includes information about the language of each document, crucial for training multilingual LLMs.

\textbf{OpenSubtitles:} This dataset~\citep{OpenSubtitles2016} consists of more than 2.6 billion sentences in more than 60 languages, obtained from subtitles of movies and TV series. To ensure data quality, the subtitles undergo preprocessing to automatically correct OCR errors, estimate the quality of each individual subtitle using metadata, and even score subtitle pairings for optimal alignment. The dataset focuses on spoken language, offering a glimpse into informal communication and slang used in movies and TV shows. 

\textbf{monoHPLT:} monoHPLT~\citep{hplt-datasets} is a collection of 75 languages from two popular sources of CommonCrawl and Internet Archive. It is collected using open-source tools which contains trillions of words. The dataset include German, French, Spanish, Italian, Greek, Dutch, Czech, Romanian, Hungarian, Swedish, Finnish, Slovak, Lithuanian, Slovenian, Estonian, and Irish languages. The monoHPLT dataset particularly focuses on low-resource languages.

\textbf{multiHPLT:} More recently, multiHPLT~\citep{hplt-datasets} is released by crawling textual data from CC and Internet Archive. It consists synthetic datasets covering 171 language pairs and 157 million sentence pairs. 

\textbf{MADLAD-400:} MADLAD-400~\citep{Madlad-400} built from CC data up to August 2022, covers 419 languages. However, to ensure data quality, it undergoes filtering, resulting in two versions: noisy and clean. The noisy version includes only basic language identification filtering, while the clean version receives more extensive filtering, though it may still contain some noise. Both versions are conveniently deduplicated and presented in a document-level format, making them well-suited for training LLMs.

\textbf{WURA:} Recently created WURA~\citep{wura} dataset  for training LLMs on multiple languages including French, Portuguese, and 16 African languages. The dataset is carefully filtered using existing multilingual datasets such mc4 to fix quality issues. Their T5 LLM trained on WURA surpass previous models on various tasks related to African languages. 

\textbf{BlBooks:} Optical Character Recognition (OCR) was used to gather multilingual datasets for British Library books. There are about 25 million pages of material on history, philosophy, geography, literature, and poetry that have been published in many languages. The majority of the content dates from the 18th and 19th centuries, while there are also some earlier works included. Books published in several European languages, primarily German and French, make up this dataset.

\textbf{ClueWeb22:} A multilingual dataset developed by the Lemur Project through crawling 10 billion web pages~\citep{Clueweb22}. It boasts several novel characteristics compared to earlier ClueWeb datasets, including a larger size, higher quality, and cleaned text. ClueWeb22 covers more than seven languages, including German, English, Spanish, French, Italian, Japanese, Dutch, Polish, Portuguese, and others.

The Culturax~\citep{culturax} is the combination of mC4 and OSCAR, is a large multilingual dataset designed for training LLMs. It boasts a massive collection of text data in 167 languages, totaling approximately 6.3 trillion tokens.

%Multilingual datasets , EU languages are the subsets of main corpora
\begin{table}[tbp]
    \centering
    \begin{tabularx}{\textwidth}{X|X}
    \hline
         \textbf{Dataset}  & \textbf{Description} \\
         \hline
         OSCAR-2301~\citep{oscar} & It is a large multilingual dataset, built from Common Crawl. The latest version is 
         OSCAR 23.01, especially good for training LLMs because it has a lot of raw web 
         data including low-resource ones.\\
         \hline
         Wikipedia~\citep{wikidump} & Wikipedia dumps, contain articles in multiple languages around 135GB.  Such articles  are divided into separate sections for each language and each section includes full articles.\\
         \hline
         CCNet~\citep{CCNet} & The dataset created from the Common Crawl covers 174 languages and uses a distinct  approach for data filtering compared to OSCAR. Large-scale removal of low-quality content, such as code or tables, results in a final compressed dataset size of around 3.2TB\\
         \hline
         mC4~\cite{mt5-mc4} & A large collection of cleaned text in 108 languages, created by analyzing data from Common Crawl. Designed specifically for pretraining LLMs. \\        
         \hline
         BlBooks & It contain 42 languages, 66GB, 28billion tokens. This dataset offers multilingual text from British library books, scanned using Optical Character Recognition (OCR) from the 18th and 19th centuries. \\
         \hline
         ROOTS~\citep{BigScience} &  A large multilingual Responsible Open-Science Open-Collaboration Text Sources (ROOTS) corpus, a 1.6TB dataset spanning 46 natural languages and 13 programming languages, was released by BigScience. The corpus is created using community-selected sources and OSCAR. \\
         \hline
         MADLAD-400~\citep{Madlad-400} & It is a large dataset consists of 3T tokens in 419 languages, built from CC. The deduplicated dataset is available in noisy version with only language identification filtering and cleaned version with more extensive filtering. \\
         OpenSubtitles~\citep{OpenSubtitles2016} & Large updated version of the OpenSubtitles, translated movies and TV subtitles. The latest version include subtitles in 60 languages, with a total of 2.6 billion sentences.\\ 
         \hline
         MonoHPLT~\citep{hplt-datasets} & Recently introduced a large dataset containing text in 75 languages. It includes commonly used languages like English, Chinese, and various European languages. The dataset consists of 5.25billion documents, 50.1TB of uncompressed text, and 5.6 trillion tokens. \\
         \hline
         WURA~\citep{wura} &  The multilingual dataset  includes textual data of French, Portuguese, and  16 African languages\\
         \hline
         ClueWeb22~\citep{Clueweb22} & A multilingual corpora developed by the Lemur Project through crawling 10 billion web pages, it covers more than seven EU languages. \\
          \hline
         W2C~\citep{w2c} & A collection of crawled corpora for 120 languages from the internet and wikipedia. \\
         \hline
         Culturax~\citep{culturax} & Combination of mC4 and OSCAR, collection of text data in 167 languages, consists of approximately 6.3 trillion tokens. \\
    \hline
    \end{tabularx}
    \caption{Pretraining Multilingual Datasets largely used to train LLMs as well as subparts of the datasets used for training monolingual foundation models. EU languages are the part these corpora.}
    \label{tab:multilingual_datasets}
\end{table}
% all the european language in OSCAR , complete statistics https://huggingface.co/datasets/oscar-corpus/OSCAR-2301

\section{LLMs for European Languages}\label{sec3:llms_4_EU_languages}
This section provides an overview of monolingual and multilingual LLMs for European languages. The primary focus of is laid on the auto-regressive models. Table \ref{tab:mono_llms} and Table \ref{tab:multilingual_llms}

\subsection{Monolingual Models}
This section provides the overview of existing monolingual pretrained small encoder-only BERT models as well as LLMs for EU languages. 

%German done
For the German language, GBERT and GELECTRA~\citep{German-llm}, were developed based on the original BERT~\citep{BERT-19} and ELECTRA~\citep{ELECTRA-llm} architectures, respectively. These models were trained on massive German text corpora including OSCAR, OPUS, Wikipedia, and OpenLegalData. The performance of both models is evaluated on downstream NLP tasks such as NER and hate speech classification. The article shows that more training data is useful and whole word masking approach improve the performance of model and  dense models outperformed smaller models. Interestingly, GELECTRA proved to be more efficient than GBERT, achieving similar performance with less training data. Later, GottBERT~\citep{GottBERT-llm}, a German LLM based on the RoBERTa~\citep{RoBERTa} architecture, was trained on large corpora. This model achieved superior performance on three downstream NLP tasks compared to previous German encoder-only models in both monolingual and multilingual setup. More recently, LeoLM~\citep{leollm} addresses the need for open-source, German-specific LLMs. It provides openly available foundation models, which serve as the backbone for various NLP tasks. The LAION institute released a 70B parameter version of LeoLM, trained with 65 billion tokens. This model is based on the Llama-2-70b architecture.

%French done
The CamemBERT~\citep{CamemBERT-french}, French LM trained on a large dataset, performs very well on various tasks like POS tagging, dependency parsing, and NER. Their model perform well using web data for training as compare to Wikipedia data,  even smaller datasets can achieve similar results to large ones. FlauBERT~\citep{flaubert-french}  is also based on orignal BERT, trained on a large multidomain French text including books, articles, web crawl data, and more. FlauBERT outperforms multilingual pretrained models on several French NLP tasks, like QA and text classification. While previous French LMs relied on encoder-only architectures, the decoder-only models include PAGnol ~\citep{PAGnol-french}, trained entirely from scratch on the massive CCNet dataset. This flexibility allows PAGnol to excel on both discriminative tasks like sentiment analysis, and generative tasks like QA and summarization. On the other hand, Cedille~\citep{Cedille-french} LLM takes a different approach by training on carefully filtered French text. It achieves competitive results on various French language tasks without any further fine-tuning. Notably, the model surpassed existing French models and even GPT-3 on zero-shot benchmarks, demonstrating its impressive capability.

\begin{table}[tbp]
    \centering
    \begin{tabular}{l|l|l|l|l|l||ll}
    \hline
        Model type & Language  & Model name  & Parameters & Tokenizer  & Tokens  & Release  \\
    \hline
        BERT/Electra & German  & BERT,ELECTRA~\citep{German-llm}  &   &  WordPiece-31K  &  & 2020\\
        RoBERTa &   & GottBERT  &   &  BPE 52K  &  & 2020\\
        Llama &   & LeoLM  & 70B &  SentencePiece-BPE  & 65B  & 2023\\
        \hline
        BERT & French  & CamemBERT~\citep{CamemBERT-french}  & 110M  & SentencePiece 32K &  65.59B  &  2020 \\
         &   & FlauBERT~\citep{flaubert-french}  & 138/373M  & BPE 50K &  65.59B  & 2020  \\
          GPT &   & Cedille~\citep{flaubert-french}  & 6B  & BPE 50K &  78.7B  & 2022  \\
           &   & PAGnol-XL~\citep{CroissantLLM-french}  & 1.5B  & BPE 50K  & 32B    & 2022  \\ 
           &   & Cedille~\citep{flaubert-french}  & 6B  & BPE 50K &  78.7B  & 2022 \\
        \hline
        GPT-2 & Italian & GePpeTto\citep{GePpeTto-italian} & 117M &  & & 2020 \\
        T5 &  & IT5-Large\citep{IT5-italian} & 738M & SentencePiece-32K &40B+ & 2022 \\
        LLaMA &   &  Fauno~\citep{Fauno-italian} & 7B, 13B   & BPE & & 2023 \\
        LLaMA &   &  Camoscio~\citep{Camoscio-italian} & 7B   & BPE & & 2023 \\
        BART &   &  BART-IT~\citep{BART-italian} &  140M  &  BPE 52K & & 2023 \\
        LLaMA-2 & & LLaMAntino~\citep{LLaMAntino-italian} & 13B & SentencePiece-BPE & 20B & 2023 \\
        \hline
        RoBERTa-L & Spanish  & MarIA~\citep{MarIA-spanish-llm-23}  &  355M   & BPE   & 135B & 2023  \\
        GPT-2-L &   & MarIA~\citep{MarIA-spanish-llm-23}   &  774M &    &  \\
        BERT &   & SPANISH-BERT~\cite{BERT-spanish-23} & 110M & SentencePiece-BPE 32K & 3B & 2023\\
        \hline
        RoBERTa & Polish  & RoBERTa-v2~\citep{Transformer-Polish}  &   &  SentencePiece-BPE & 15-30B  & 2020  \\
        BERT & & HerBERT~\citep{HerBERT-Polish} & 110-340M & BPE & & 2021 \\
        mT5 & & plT5~\citep{KLEJ-Polish} & 820M & sentencepiece-50K & & 2022 \\
         Mistral-7B & & Curie-7B-v1~\citep{polish-llms} & 7.24B & BPE & 276M & 2024 \\
        \hline
        BERT & Romanian  & RoBERT~\citep{Romanian-BERT-20}  & 341M & WordPiece   &  2.07B  & 2020  \\
        GPT-2 &   & RoGPT-2~\cite{Romanian-GPT2-21}   & 774M  &  BPE & 2.56B  &  2021 \\
        AlBERT~\citep{albert-20} &   & ALR-BERT~\citep{lite-romanian-bert-22}   &   &  BPE-50K &  2.42B &  2022\\
        \hline
        RoBERT & Dutch & RobBERT~\citep{Dutch-RoBERTa}  &  117M  &  BPE & 6.6B & 2020 \\
         GPT &  & GPT-NEO1.3b~\citep{gptneo-dutch}  &  1.3B  &  BPE & 33B & 2022 \\
         LLaMA-2  &  & llama2-13b~\citep{dutch-lrs-23}  & 13B    &   & 2B & 2023  \\      
        \hline
        BERT &   &  Greek-BERT~\citep{GREEK-BERT20} &  110M & BPE-35K   & 3.04B  & 2020 \\
        BART~\citep{BART22} & Greek  & GreekBART~\citep{GreekBART23}  & 181M  &  SentencePiece-50K  &  & 2023 \\
        T5 &   &  GreekT5~\citep{GreekT5-23} & 580M  &    &  & 2023 \\
        \hline  
       GPT-3 &  Hungarian & PULI~\citep{Hungarian-gpt23}  &  6.7B  &    & 41.50B & 2023  \\  
        \hline  
        BERT & Swedish  & KB-BERT~\citep{swedish-bert20}  &  3.497B & SentencePiece50   & & 2020   \\
        GPT3 &   & GPT-SW3~\citep{GPT-swedish22}  & 3.5B  & SentencePiece   & & 2023   \\
        CTRL &   &  SweCTRL-Mini~\citep{SwedishCTRL-mini23}  &   1.63B & BPE & 5.743B    & 2023 \\
        \hline
        ALBERT & Czech & Czech-ALBERT~\citep{Albert-Czech-2020}  & & SentencePiece & 10B & 2020\\
        BERT-ALBERT &  & Czert-BERT~\citep{BERT-Czech-2021}  & 109M  & WordPiece-30K &  & 2023 \\
        RoBERTa &  & Robe-BERT~\citep{RobeCzech-21}  & 125M  & BPE-52K  &  4.917B  & 2021  \\
        \hline        
        BERT & Portuguese  & BERT~\citep{BERT-Portuguese-23}  & & WordPiece30K  &  2.7B  & 2023   \\
        & & BERTabaporu~\citep{BERTabaporu-portuguese23} & &335M & 2.9B & 2023\\
        GPT-J, LLaMA &   & Sabia~\citep{Portuguese-llm-23}  && 65B  & 7.8B & 2023 \\
        LLaMA &   & Cabrita~\citep{Cabrita-portuguese23}  & &3B &  7B & 2023   \\ 
        LLaMA-2 &   & Bode~\citep{Portuguese-bode-llm24}  & &7-13B &  7B  & 2024   \\
        \hline
        GPT-2 & Bulgarian  & GPT-WEB-BG~\citep{GPT-WEB-BG23}  & 1.5B  & & & 2023 \\
        \hline
        BERT & Finnish  & FinBERT~\citep{bert-finnish-19} & 110M  &  ~\citep{swedish-bert20}BPE-50K & 3.3B & 2019   \\ 
        GPT &   & FinnGPT~\citep{gpt-finnish-22}  & 1B  &  BPE-50K &  & 2022   \\ 
        GPT-2 &  & FinnishGPT2~\citep{gpt2-finnish}  & 0.78B  &  BPE-50K &  & 2022   \\
        GPT-2 &  & FinGPT~\citep{FinGPT}  & 186M-13B  &  BPE & 38B & 2023 \\
         \hline
        BERT & Slovak  & SlovakBERT~\citep{SlovakBERT22}  & 125M  &  BPE-50K & 4.6B & 2022  \\
        \hline
       T5 & Slovene  & SloT5~\citep{Slovenian-T5-23}  &  750M & & 4.20B & 2023 \\
       \hline
        BERT & Estonian  &  EstBERT~\citep{EstBERT-estonian} & 110M &  & 2021 \\
          \hline
          BERT & Maltese   & BERT~\citep{BERT-Maltese-22} &   & & 46.66B &2022 \\
    \hline
    \end{tabular}
    \caption{An overview of the existing work on official European languages general pretrained Transformer based small and large language models. Few recent survey papers can be referred for the existing work on English\citep{min2023recent, minaee2024large}. The languages are ordered in terms of number of native speakers. }. 
    \label{tab:mono_llms}
\end{table}

%Italian  done
Significant progress has been made in developing LLMs for the Italian language. Fauno~\citep{Fauno-italian}, an Italian LLM, is trained on a unique dataset of synthetic data. This includes English-to-Italian translations of text from social media platforms like Quora and Stack Overflow, as well as the Alpaca dataset. Built on the LLaMA model, Fauno is further fine-tuned with Italian conversation datasets translated from diverse sources, encompassing medical data, technical content, and social media dialogues. Another Italian LLM, GePpeTto~\citep{GePpeTto-italian}, leverages the GPT-2 architecture. Its evaluation involved both automatic metrics and human judgment. Automatic evaluation assessed GePpeTto's performance across different writing styles and its sentence structure, while human evaluation focused on a sentence completion task. The study revealed that GePpeTto's generated sentences closely resembled human-written ones. Camoscio~\citep{Camoscio-italian} stands out as an Italian LLM specifically designed to follow user instructions. It is based on a fine-tuned LLaMA-7B model using an instruction-tuning dataset derived from an Italian translation of the Stanford Alpaca dataset. Unlike previous models, Camoscio is freely available for researchers. Focusing on text summarization, BART-IT~\citep{BART-italian} is an Italian LLM built upon the BART~\citep{BART22} architecture. This model surpasses existing options in Italian text summarization tasks. More recently, LLaMAntino~\citep{LLaMAntino-italian} was developed specifically for Italian by adapting the LLaMA-2 model. Fine-tuned for the Italian language, LLaMAntino significantly enhances the model's understanding and generation of Italian text. 

%Spanish done
A BERT model specifically pretrained for Spanish, as described in~\citep{BERT-spanish-23}, leverages text from OPUS and Wikipedia. To evaluate this Spanish BERT, researchers proposed the Spanish Benchmark, a collection of Spanish-specific NLP tasks similar to the GLUE benchmark~\citep{2018glue} for English. The model outperforms previous multilingual models. In contrast, MarIA~\citep{MarIA-spanish-llm-23} offers a variety of pretrained models, including both base and large versions of RoBERTa and GPT-2. All these models are trained on a large dataset from the Spanish National Library. Consequently, MarIA models surpass previous Spanish LLMs on various tasks.

%Polish done
For Polish, two Transformer-based LMs~\citep{Transformer-Polish}leverage the BERT architecture and are trained on a massive dataset of 1 billion Polish sentences, corresponding to approximately 135GB of raw text. The proposed LMs demonstrate significant improvements on thirteen Polish language tasks, outperforming existing approaches on eleven of them. Building upon this foundation, HerBERT~\citep{HerBERT-Polish} specifically designed for the Polish language. HerBERT leverages knowledge transfer from pretrained multilingual LMs to enhance the performance of Polish BERT models. When evaluated on eleven downstream NLP tasks, HerBERT outperforms previous models on eight tasks, achieving the top position on the KLEJ~\citep{KLEJ-Polish} Benchmark. Moreover, plT5~\citep{KLEJ-Polish} is a encoder-decoder (text-to-text) model for Polish. It leverages unsupervised denoising pretraining, where the model weights are initialized with mT5~\citep{mt5-mc4} model. The plT5 achieves better performance than decoder-only models on tasks including the KLEJ Benchmark, MT, and QA. More recently,~\citep{polish-llms} introduce a new pre-training technique called Language Adaptive Pre-training (LAPT) to train the Curie-7B-v1 model for Polish. This decoder-only model performs competitively on eight downstream tasks. Notably, unlike traditional encoder-decoder models, Curie-7B-v1 can both predict masked tokens and generate high-quality Polish text. 
 
%Lithuanian no work 

%Finnish done
Earlier work on Finnish language modelling based on Transformer models can be traced back to BERT~\citep{bert-finnish-19} trained on 3.3billion tokens consists of 110 parameters was evaluated on the text classification tasks such as POS tagging and NER. The pretraining data include Yle corpus, STT corpus, news articles, Suomi24, and Common Crawl. Later FinnGPT~\citep{gpt-finnish-22} explores the effectiveness GPT for Finnish language by introducing three LLMs consists of 0.13B, 0.25B, and 01B parameters. All the models were evaluated on various tasks including text classification, language modelling, text generation, and sentiment analysis. The largest FinnGPT-1B outperform small models by generating relevant and good quality text. Later, GPT-2~\citep{gpt2-finnish} model pretrained on a  Finnish text data consists of 84GB data in a self-supervised fashion using Finnish subset of the mC4, wikipedia, Yle, news archive (STT), and Suomi24 datasets. More recently, FinGPT~\citep{FinGPT} is released by training GPT model on large Finnish corpora by combining web crawls, news articles, social media content, and eBooks. Overall, seven Finnish-only models trained from scratch ranging in size from 186M to 13B parameters. Moreover, multilingual BLOOM~\citep{BLOOM-22} model is enhanced by continual pretraining with a combination of its original data and Finnish corpora. 

%Romanian low resource done
The RoBERT~\citep{Romanian-BERT-20} model for Romanian, based on the BERT architecture, outperforms existing models on seven different NLP tasks related to sentiment analysis, dialect identification, and diacritics restoration. As shown by a comparison with other multilingual and Romanian-only BERT models, RoBERT achieved the best performance on all seven tasks.  Masala et al.~\citep{lite-romanian-bert-22} propose an ALR-BERT model trained on Romanian text. This model combines Romanian Wikipedia entries with OSCAR and OPUS corpora. The performance on various tasks, including POS tagging, and compare it to existing Romanian and multilingual BERT models. Another noteworthy contribution is Romanian GPT-2~\citep{Romanian-GPT2-21}. This model comes in three versions (base, small, and large) and is trained on a corpus of Romanian text including OSCAR, Wikipedia, books, news, and dialogues. RoGPT2 performs well on various tasks and surpasses existing automatic grammar correction models.

%Dutch done
A Dutch language model, namely RobBERT~\citep{Dutch-RoBERTa}, is based on the RoBERTa architecture. This model excels at handling limited data, and its specifically designed tokenizer demonstrates good performance. It is pre-trained on the Dutch portion of the OSCAR corpus and performs well on various downstream NLP tasks, including coreference resolution, POS tagging, and NER. More recently, ~\citep{dutch-bert-24} improves upon prior RobBERT~\citep{Dutch-RoBERTa} models by using pretrained weights from RoBERTa instead of training a foundation model.
To handle the new Dutch tokenizer, a technique called "token translation" is used to adapt the existing vocabulary. Their results show that RobBERT-2023 outperforms its predecessor on various NLP tasks while requiring less training time. Similar to~\citep{Dutch-RoBERTa}, the model is trained on dutch subset of OSCAR corpus. Moreover,~\citep{dutch-lrs-23} introduces Dutch LLMs, language resources, translation repository, datasets and a benchmark leaderboard. Two fine-tuned variants Llama 2-based LLMs, four synthetic datasets for instruction following, translation, and a leaderboard for Dutch generative model benchmarks.

%Greek  done
For Greek,~\citep{GREEK-BERT20} proposed a monolingual model named GREEK-BERT, similar to the original BERT~\citep{BERT-19} architecture. Trained on a 29GB dataset, GREEK-BERT outperforms previous models in tasks like POS tagging, NER, and relationship extraction. Building upon this success, ~\citep{GreekBART23} developed GreekBART, an encoder-decoder model based on the BART~\citep{BART22}. This model specifically focuses on text generation in Greek.  GreekBART surpasses previous models in tasks like summarization, and the authors also introduced a new GreekSUM dataset for evaluation. Continuing the trend in summarization, ~\citep{GreekT5-23} released GreekT5, a series of models utilizing pretrained models from Hugging Face for summarizing Greek news articles. Inspired by T5, their approach demonstrates that even smaller models can achieve performance comparable to denser models.

%Hungarian done
The early work on Transformer-based language modeling in Hungarian can be traced back to huBERT~\citep{huBERT20}. The huBERT model is trained on the Hungarian part of Wikipedia and Webcorpus 2.0. It outperforms mBERT on several Hungarian benchmark tasks. A recent work presented by~\citep{Hungarian-gpt23} released a GPT-3 based monolingual model named PULI for the Hungarian language, as well as multilingual models trained on Hungarian, English, and Chinese languages, named PULI-GPTrio, with more than 1TB of text data. The PULI-GPTrio was fine-tuned with the Alpaca instruction dataset.

%Swedish done
Recent advancements in Swedish LMs have yielded promising results. Swedish KB-BERT~\citep{swedish-bert20} achieves  the SOTA performance on various downstream NLP tasks. Notably, their training corpus, curated from diverse web sources, consists of a massive 3.497 billion words. Later, GPT-SW3~\citep{GPT-swedish22} is decoder-only model specifically designed for Swedish. The model was trained on a large corpus of more than 100GB of text and boasts 3.5 billion parameters. Their findings demonstrate that GPT-SW3 outperforms other models of similar size. More recently, researchers released SweCTRL-Mini~\citep{SwedishCTRL-mini23}, a Swedish LLM built on the CTRL model~\citep{ctrl-llm19} for user-controlled text generation. While SweCTRL-Mini achieves performance comparable to GPT-SW3, the authors acknowledge some limitations compared to GPT-3.

%Bulgarian done
Little work exists in Bulgarian language modeling. Recently, two models, an encoder-only BERT-WEB-BG and a decoder-only GPT-WEB-BG, were released for the Bulgarian language~\citep{GPT-WEB-BG23}. Both models are trained on a newly crawled web dataset. These models perform well in named entity recognition (NER) tasks and other text classification tasks. 
%Czech done
For Czech, researchers have primarily focused on encoder-only models. Earlier work on developing a Czech ALBERT model~\citep{Albert-Czech-2020} demonstrated the best performance in text classification tasks, but lower performance in QA. Afterwards, Czert~\citep{BERT-Czech-2021}, a BERT and ALBERT-based model for the Czech language, outperformed multilingual models on nine downstream NLP tasks. Later, RobeCzech~\citep{RobeCzech-21}, specifically trained for Czech and based on the RoBERTa architecture, surpassed previous Czech language models. This includes multilingual models and other Czech-specific models like Czert and ALBERT, achieving better results on five NLP tasks.

%Portuguese done
Recent advancements have led to a rapid catch-up for Portuguese LLMs, with encoder-only models reaching GPT-3 architectures. BERTimbau~\citep{BERT-Portuguese-23}, a BERT-based model, outperformed previous models on sentence similarity and NER tasks. This model leverages the BERT architecture and is trained on the brWaC corpus~\citep{brWaC-corpus}. Subsequently, BERTabaporu~\citep{BERTabaporu-portuguese23}, a new LLM specifically trained for Brazilian Portuguese using a Twitter dataset is built on the BERT model. It surpasses existing general-purpose Portuguese LMs in text classification tasks. More recently, Sabia~\citep{Portuguese-llm-23}, presents a family of LLMs trained on Portuguese text by using popular GPT-J~\citep{gpt-j} and LLaMA~\citep{LLaMA} transformer architectures. Notably, Sabia outperform previous monolingual and multilingual models on various Portuguese NLP tasks, even with less data compared to standard training. However, their performance weakens on English tasks. The openCabrita models, introduced in~\citep{Cabrita-portuguese23}, utilize continuous pretraining methods, resulting in a 3-billion-parameter OpenLLaMA model trained solely on Portuguese. A filtered Portuguese subset of the large public mC4 dataset is used for training. The openCabrita-3B model employs a special tokenizer that significantly reduces the number of tokens needed, yet maintains performance comparable to traditional methods. Building on this progress, Bode~\citep{Portuguese-bode-llm24}, a recently released fine-tuned LLaMA-2 model specifically for Portuguese, was trained on a Portuguese subset of Alpaca. This fine-tuning enables Bode to understand and generate Portuguese text effectively, performing well in downstream NLP tasks like sentiment analysis and news categorization. Most recently, GlórIA~\citep{Portuguese-gpt-24}, a Portuguese LLM trained on a massive dataset of 35 billion words from various sources, has been released. Additionally, a benchmark was established to evaluate Glória's capabilities. Compared to previous Portuguese LLMs, Glória demonstrates significantly better performance in language understanding and generation. 

%low resource languages
Several EU languages, including Maltese, Lithuanian, Slovak, Slovene, Irish, Latvian, Estonian, and Croatian, are considered low-resource languages for NLP tasks. This means there is a limited amount of online textual resources and annotated data available for training LLMs in these languages. Although these languages are often included in multilingual pretrained models, they typically lack dedicated LLMs trained specifically for them.

SlovakBERT \citep{SlovakBERT22} achieves state-of-the-art (SOTA) performance on various downstream tasks, including POS tagging and sentiment analysis. Additionally, the article introduces a benchmarking dataset.

SloBERTa \citep{SloBERTa-21} introduces the first BERT based model for the Slovene language, trained on masked language tasks. Comparisons of SloBERTa with existing multilingual models show that SloBERTa performs better in multiple tasks, outperforming these models in POS tagging, NER, sentiment analysis, and word analogy tasks.

T5 models \citep{Slovenian-T5-23} are released for the Slovenian language and compares them with existing BERT models for text classification and generation tasks. Slovenian-BERT performs better for classification tasks, while Slovenian-T5 shows promising results for text generation tasks.

gaBERT and gaELECTRA \citep{gaBERT-irish} are monolingual BERT and ELECTRA models for the Irish language, trained on several datasets, including the Irish dependency parsing dataset CoNLL-17, the New Corpus for Ireland (NCI), and Irish subsets of OSCAR, ParaCrawl, and Wikipedia.

LVBERT \citep{latvian-bert} is a small Latvian-specific BERT model that outperforms non-contextual representations and the multilingual BERT model on various downstream Latvian NLP tasks.

EstBERT \citep{EstBERT-estonian} presents a BERT model for Estonian language modeling. It outperforms the multilingual BERT in five out of seven NLP tasks, including POS tagging, NER, dependency parsing, and text classification.

%multiple monolingual BERT model trained using wiki ddata
% ~\citep{BERT-19} Initially, BERT as trained on English and Chinese languages, and then shortly release multilingual BERT (mBERT) capable of handling 104 languages. The BERT relies on encoders- decoders transformer architecture but at the pretraining stage for mBERT, only encoders are used. The mBERT is trained on a large textual data 104 languages obtained form various sources such as Wikipedia or web crawl. However, multilingualism do not guarantee the equal performance across all languages. It heavily depends on the few factors including the amount of training data. 

% ~\citep{WikiBERT} presents WikiBERT, mono-lingual BERT models for 42 languages using Wikipedia data. The WikiBERT also include European languages and the evaluation demonstrate that most of the monolingual WikiBERT models surpass a general multilingual BERT (mBERT) in a close test in predicting missing words. 

\subsection{Multilingual language models} 
While most LLMs presented in section ~\ref{sec2:Background}, have been trained on multilingual corpora~\citep{mt5-mc4,wikidump,Madlad-400,OPUS,OpenSubtitles2016,Roots}, their primary focus for training and evaluation has remained on English. In this section, we present an overview of bilingual and multilingual LLMs trained and evaluated on European languages.
\begin{table}[!h]
    \centering
    \begin{tabular}{l|l|l|l|l|l||ll}
    \hline
        Model type & Languages  & Model name  & Parameters & Tokenizer  & \#Tokens  & Release  \\
        \hline
        GPT & en, hu, cn  & PULI-GPTrio~\citep{Hungarian-gpt23} & 6.7 &   & 1078GB  & 2023  \\
        \hline
        Llama & fr, en  & CroissantLLM\citep{CroissantLLM-french} &   1.3B & SentencePieceBPE  & 3T  & 2024  \\
        \hline
        Bloom & en, fi, code  & Poro34B~\citep{poro34b-24} & 34B  &  BPE & 129B & 2024  \\
        \hline
        Megatron-LM & fi \& others  & Aurora-M~\citep{aurora-m24}  &  15B &   & 377B & 2024 \\
        Bloom & Swedish and 6 others  & Viking\footnote{\url{https://www.silo.ai/blog/viking-7b-the-first-open-llm-for-the-nordic-languages}}  & 7B,13B,33B &   & & 2024 \\
    \hline
    \end{tabular}
    \caption{An overview of the multilingual LLMs for official European  languages. Few recent survey papers on English \citep{min2023recent, minaee2024large, zhao2024explainability}}
    \label{tab:multilingual_llms}
\end{table}
% LLaMA &  French-English & CroissantLLM~\citep{CroissantLLM-french}  & 1.3B  & LlamaTokenizer 32K  & 303B    & 2024  \\ 
There has been a little efforts in the development of multilingual LLMs capable of handling multiple European languages effectively. One such example is CroissantLLM~\citep{CroissantLLM-french}, a French-English bilingual LLM trained on a massive dataset with an equal mix of French and English data. This model also proposes a new benchmark specifically designed to evaluate the performance of French language models.

Another recent development is Poro34B~\citep{poro34b-24}, a decoder-only model trained on Finnish, English, and even programming languages. Similar to models like FinGPT~\citep{FinGPT} and BLOOM~\citep{BLOOM-22}, Poro34B boasts improvements for better training efficiency. This has resulted in Poro34B significantly outperforming previous Finnish language models in translation tasks and generating high-quality English and code.

Continuing this trend, the AURORA-M model~\citep{aurora-m24} is a recently released open-source multilingual LLM. This 15 billion parameter model tackles six languages: Finnish, English, Japanese, Hindi, Vietnamese, and code. AURORA-M utilizes continual pretraining on general, multilingual web data and is evaluated on various languages and tasks through benchmarks like FIN-bench. Notably, AURORA-M demonstrates strong performance in multilingual scenarios while remaining competitive in English and code generation.

Most recently, the TurkuNLP group and Silo-AI released the Viking models~\citep{viking}. These models come in different sizes (7B, 13B, and 33B parameters) and prioritize excellence in handling Nordic languages. Building upon Poro's foundation designed for low-resource languages, Viking expands its capabilities to encompass Danish, Swedish, Norwegian, and Icelandic. While Viking shares the same training setup as Poro, it demonstrates a particular strength in handling low-resource languages.

% \section{Discussion}\label{sec12}
% Discussions should be brief and focused. In some disciplines use of Discussion or `Conclusion' is interchangeable. 

\section{Conclusion}\label{sec13}
This paper provides an overview of various LLM families, including LLaMA, PaLM, GPT, and MoE. We discuss the methods used to create and enhance LLMs for official European languages, focusing on reviewing existing work on LLM development for these languages. To the best of our knowledge, this is the first comprehensive review of resources and development methods for LLMs in this domain. Our key contributions include a summary of the available monolingual and multilingual datasets for training LLMs in EU languages, as well as background information, pretraining datasets, and future research directions.

% \section*{Declarations}

% Some journals require declarations to be submitted in a standardised format. Please check the Instructions for Authors of the journal to which you are submitting to see if you need to complete this section. If yes, your manuscript must contain the following sections under the heading `Declarations':

% \begin{itemize}
% \item Funding
% \item Conflict of interest/Competing interests (check journal-specific guidelines for which heading to use)
% \item Ethics approval and consent to participate
% \item Consent for publication
% \item Data availability 
% \item Materials availability
% \item Code availability 
% \item Author contribution
% \end{itemize}

%\bibliographystyle{plainnat}
\bibliography{article}

\end{document}